\documentclass{article}

\usepackage{microtype}
\usepackage{graphicx}
\usepackage{subfigure}
\usepackage{booktabs} 
\usepackage{amsmath}
\usepackage{amsfonts}

\usepackage[hyphens]{url}
\usepackage{hyperref}
\hypersetup{colorlinks=true,breaklinks=true}
\usepackage[accepted]{icml2021}

\icmltitlerunning{Emo3D: Metric and Benchmarking Dataset for 3D Facial Expression Generation from Emotion Description}

\begin{document}

\twocolumn[
\icmltitle{Emo3D: Metric and Benchmarking Dataset for 3D Facial Expression Generation from Emotion Description}




\icmlsetsymbol{equal}{*}

\begin{icmlauthorlist}
\icmlauthor{Mahshid Dehghani}{dh,cs}
\icmlauthor{Amirahmad Shafiee}{equal,dh,cs}
\icmlauthor{Ali Shafiei}{equal,dh,cs}
\icmlauthor{Neda Fallah}{dh,cs}
\icmlauthor{Farahmand Alizadeh}{dh,ee}
\icmlauthor{Mohammad Mehdi Gholinejad}{dh,ce}
\icmlauthor{Hamid Behroozi}{ee}
\icmlauthor{Jafar Habibi}{ce}
\icmlauthor{Ehsaneddin Asgari}{qatar}

\end{icmlauthorlist}

\icmlaffiliation{dh}{NLP \& DH Lab, Department of Computer Engineering, Sharif University of Technology, Tehran, Iran}

\icmlaffiliation{cs}{Department of Mathematical Sciences, Sharif University of Technology}

\icmlaffiliation{ee}{Department of Electrical Engineering, Sharif University of Technology}

\icmlaffiliation{ce}{Department of Computer Engineering, Sharif University of Technology}

\icmlaffiliation{qatar}{Qatar Computing Research Institute, Doha, Qatar}

\icmlcorrespondingauthor{Ehsaneddin Asgari}{\mbox{easgari@hbku.edu.qa}}

\icmlkeywords{Machine Learning, ICML}

\vskip 0.3in
]



\printAffiliationsAndNotice{\icmlEqualContribution} 

\begin{abstract}
  Existing 3D facial emotion modeling have been constrained by limited emotion classes and insufficient datasets. This paper introduces ``Emo3D'', an extensive ``Text-Image-Expression dataset'' spanning a wide spectrum of human emotions, each paired with images and 3D blendshapes. Leveraging Large Language Models (LLMs), we generate a diverse array of textual descriptions, facilitating the capture of a broad spectrum of emotional expressions. Using this unique dataset, we conduct a comprehensive evaluation of language-based models' fine-tuning and vision-language models like Contranstive Language Image Pretraining (CLIP) for 3D facial expression synthesis. We also introduce a new evaluation metric for this task to more directly measure the conveyed emotion. Our new evaluation metric, Emo3D, demonstrates its superiority over Mean Squared Error (MSE) metrics in assessing visual-text alignment and semantic richness in 3D facial expressions associated with human emotions. ``Emo3D''  has great applications in animation design, virtual reality, and emotional human-computer interaction.
\end{abstract}

\section{Introduction}
Automatic translation of character emotions into 3D facial expressions is an important task in digital media, owing to its potential to enhance user experience and realism. Facial Expression Generation (FEG) has a wide range of applications across various industries, including game development, animation, film production, and virtual reality. Previous studies in this domain have primarily focused on generating facial expressions for 2D or 3D characters, often relying on a limited set of predefined classes~\cite{siddiqui2022explore} or driven by audio cues~\cite{Karras2017, Peng2023}. However, there is a growing need for better control in the generation of complex and diverse human facial expressions. Recent studies~\cite{Zou2023, Zhong2023, Ma2023} have made notable progress in this area through the use of text prompts, offering a more direct approach to address the challenge of limited control that has been prevalent in earlier works~\cite{siddiqui2022explore, Karras2017, Peng2023}.

The primary issue with recent works using text prompts is (i) their limited focus on textual descriptions of emotions. Many studies have not deeply explored emotional context. These studies have not offered a comprehensive solution that integrates both textual descriptions and 3D FEG, creating a noticeable gap in the field~\cite{Zhong2023, Zou2023}. Moreover, there is (ii) a scarcity of datasets containing emotional text alongside corresponding 3D facial expressions, impeding the development and training of FEG models for practical applications~\cite{Zhong2023, Zou2023, Ma2023}. Additionally, (iii) the absence of reliable benchmarks and standardized evaluation metrics in this research area further complicates the assessment of FEG models.\\
\noindent\textbf{Contributions:} This paper tackles key challenges in FEG, focusing on generating expressions from textual emotion descriptions. Our contributions towards addressing the gaps in the field of FEG are as follows:
\noindent\textbf{(i) Emo3d-dataset:} We present the Emo3D-dataset, specifically developed to bridge the gap between textual emotion descriptions and 3D FEG. This dataset provides a rich compilation of annotated emotional texts alongside matching 3D expressions for effective training and assessment of FEG models. \\
\noindent\textbf{(ii) Baseline Models:} We propose baseline FEG models as benchmarks for future research in this field. These models offer a reference point for evaluating new advancements and assessing progress.\\ \noindent\textbf{(iii) Evaluation Metric:} To address the absence of standardized evaluation metrics in FEG, we introduce a new metric designed for the unique challenges of capturing the complexities of human emotions.
\section{Related Work}
\noindent\textbf{Audio-based emotion extraction:} FEG methods often utilize audio data, leveraging the semantic, tonal, and expressive qualities of voice for 3D generation. ``Audio-driven Facial Animation''~\cite{Karras2017} learns to map audio waveforms to 3D facial coordinates, identifying a latent code for expression variations beyond audio cues. ``EmoTalk''~\cite{Peng2023} focuses on creating 3D facial animations driven by speech, aligning expressions with both content and emotion.

\noindent\textbf{CLIP-based baselines:} The utility of CLIP's language-and-vision feature space~\cite{Radford2021} in text-to-image generation has been highlighted in several works. MotionCLIP~\cite{Tevet2022} leverages CLIP for a feature space that accommodates dual modalities, enabling out-of-domain actions and motion integration into CLIP's latent space. The 4D Facial Expression Diffusion Model~\cite{Zou2023} introduce a generative framework for creating 3D facial expression sequences, utilizing a Denoising Diffusion Probabilistic Model (DDPM). The framework consists of two tasks: learning a generative model based on 3D landmark sequences and generating 3D mesh sequences from an input facial mesh driven by the generated landmarks. Also, ExpCLIP~\cite{Zhong2023} is an autoencoder designed to establish semantic alignment among text, facial expressions, and facial images. ExpClip introduces a blendshape encoder to map blendshape weights to an embedding, reconstructed by a decoder. Concurrently, a CLIP text encoder $(\epsilon_{\text{text}})$ and text projector $(P_{\text{text}})$, along with an image encoder $(\epsilon_{\text{img}})$ and an image projector $(P_{\text{img}})$ to map emotion text and images into a joint embedding space.\\
Additionally,~\cite{Li2023} introduced CLIPER, a unified framework for both static and dynamic facial expression recognition, utilizing CLIP and introducing multiple expression text descriptors (METD) for fine-grained expression representations, achieving state-of-the-art performance by a two-stage training paradigm which involves learning METD and fine-tuning the image encoder for discriminative features.

\noindent\textbf{Metrics:}While a variety of metrics exist for evaluating 2D image generation, the development of effective metrics for 3D FEG remains a challenge. Building upon the approach in ~\cite{Xu2017}, ~\cite{Cong2023} adopted R-precision to measure the alignment between input text and output image. This metric was calculated using a CLIP model fine-tuned on the entire dataset, following the methodology outlined in ~\cite{Park2021}.
\section{Dataset}
\begin{figure}[ht!]
     \centering
        {\includegraphics[width=0.5\textwidth, keepaspectratio]{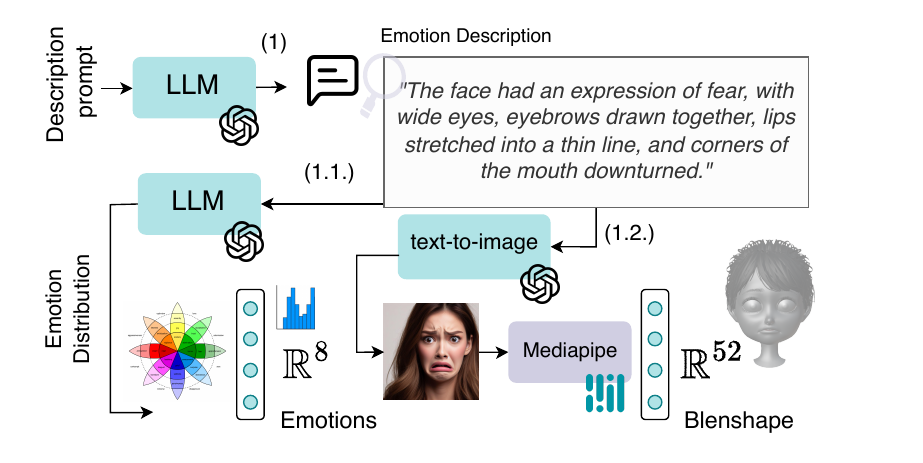}}
        \caption{Emo3D Dataset Creation: Textual data describing human emotions is initially generated using GPT \cite{OpenAI2023}. We then utilize DALL-E models \cite{ramesh2022hierarchical} to synthesize human faces. Each image undergoes face blendshape extraction using MediaPipe \cite{Lugaresi2019}. Furthermore, we employ GPT \cite{OpenAI2023} to extract the emotion distribution for each prompt.}
        \label{fig:architecture}
\end{figure}

We introduce the Emo3d-dataset, an assembly of 150,000 instances. Each instance comprises a triad: textual description, corresponding image, and blendshape scores created as follows: 

\noindent\textbf{(i) Emotion descriptions:} 
To generate emotion-specific textual descriptions, we prompted GPT-3.5 ~\cite{OpenAI2023} to focus on eight primary emotions: happiness, anger, surprise, sadness, disgust, contempt, fear, and neutral. Subsequently, we again utilized GPT-3.5 to derive emotion distributions for these textual elements through carefully crafted prompts. This process resulted in eight-dimensional vectors representing distinct emotional profiles, as illustrated in Figure \ref{fig:architecture}. A more in-depth analysis of the linguistic characteristics of the generated data can be found in Appendix \ref{sec.Analysis}.

\noindent\textbf{(ii) 2D Image Generation:} Subsequently, we utilize DALL-E 3~\cite{ramesh2022hierarchical}, an image generation model, to create images that align with the generated textual descriptions. A detailed view can be found in Appendix \ref{sec.Analysis}.

\noindent\textbf{(iii) Blendshape scores estimation:} We employ Mediapipe framework~\cite{Lugaresi2019} to synthesize blendshape scores corresponding to the generated images based on textual descriptions. 

\begin{figure}[htbp!]
     \centering
        {\includegraphics[width=0.35\textwidth, height=0.35\textwidth]{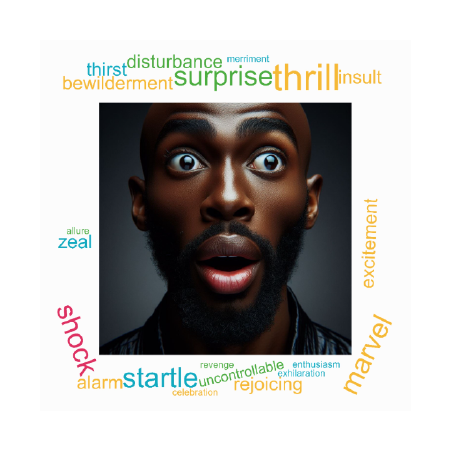}}
        \vspace{-6 mm}
         \caption{``Surprise'' Emotion Word Cloud: closest words to ``surprise'' using Emolex~\cite{LREC18-AIL} based on cosine similarity of emotion distribution.}
        \label{fig:surprise}
\end{figure}

\noindent\textbf{Primitive Emotion Faces:} Additionally, for intrinsic evaluation purposes, we construct a dataset of primitive emotions comprising singular emotion words, each paired with corresponding images that portray males and females exhibiting three distinct intensity levels of emotion. Utilizing Mediapipe~\cite{Lugaresi2019}, we subsequently extract blendshape scores for the facial expressions depicted in these images. The emotional distributions associated with these individual words are derived using Emolex~\cite{LREC18-AIL}. An example is provided in Figure \ref{fig:surprise}.\\

\noindent\textbf{Human Evaluation:}
Human evaluation was conducted to assess the accuracy of the emotion distribution of the dataset. The results showed a high agreement rate of $81\%$, indicating that the generated emotion distributions are mostly aligned with human understanding of emotions.
The human evaluation is described in more detail in the Appendix \ref{sec.human}.\\

\noindent\textbf{Comparison of Emo3D with existing datasets:} Emo3D stands out for its integration of textual, visual, and blendshape modalities, offering a comprehensive representation of emotional expressions. Compared to single-modal datasets~\cite{saravia-etal-2018-carer,Mollahosseini_2019,8e31-3188-23}, Emo3D provides a more holistic representation of emotional expressions. Additionally, With 90,000 images and 60,000 texts, it facilitates emotion recognition in both modalities. Emo3D's emphasis on emotion descriptions distinguishes it from other datasets like TEAD~\cite{Zhong2023} and TA-MEAD\cite{Ma2023}, which focus on different aspects of FEG. Furthermore, it can be employed for emotion recognition in text and images, thanks to the associated emotion distributions. Table~\ref{tab:dataset-comparison} presents a comparison of Emo3d with existing datasets.

\begin{table}[htbp!]
\centering
\setlength{\tabcolsep}{0.4pt} 
\begin{tabular}{@{}lccc@{}}
\toprule
\textbf{Dataset} & \textbf{Size} & \hspace{-15pt} \textbf{Emotion Distribution} & \textbf{Modalities} \\
\midrule
AffectNet & 440,000 & Yes & Image \\
\midrule
Emo135 & 700,000 & Yes & Image \\
\midrule
CARER & 417,000 & Yes & Text \\
\midrule
TEAD & 50,000 & No & \begin{tabular}[c]{@{}c@{}}Text\\ Blendshape\end{tabular} \\
\midrule
TA-MEAD & - & No & \begin{tabular}[c]{@{}c@{}}Text\\ Video\end{tabular} \\
\midrule
Emo3D & \begin{tabular}[c]{@{}c@{}}Text: 60,000\\ Image: 90,000\\ Triple: 150,000\end{tabular} & Yes & \begin{tabular}[c]{@{}c@{}}Text\\ Image\\ Blendshape\end{tabular} \\
\bottomrule
\end{tabular}
\caption{Comparison of Emo3D with existing datasets:This table summarizes key attributes of Emo3D alongside established datasets such as AffectNet~\cite{Mollahosseini_2019}, Emo135~\cite{8e31-3188-23}, CARER~\cite{saravia-etal-2018-carer}, TEAD~\cite{Zhong2023}, and TA-MEAD~\cite{Ma2023}.}
\label{tab:dataset-comparison}
\end{table}

The Emo3d-dataset shares similarities with other existing datasets, particularly TEAD\cite{Zhong2023} and TA-MEAD\cite{Ma2023}, in terms of modality integration and a focus on emotional expressions.

The TA-MEAD~\cite{Ma2023} dataset, designed for 2D FEG , provides emotion descriptions for videos, along with Action Unit (AU)~\cite{ekman1978facial} intensity annotations for each video. In contrast, our Emo3d-dataset offers a unique perspective by concentrating on textual emotion expressions, corresponding images, and blendshape scores.

The TEAD~\cite{Zhong2023} dataset, designed for 3D FEG, features situation descriptions, our Emo3d-dataset distinguishes itself by emphasizing emotion descriptions. Additionally, our dataset includes a distinctive feature with corresponding images for each text, providing a richer and more comprehensive resource. The Emo3d-dataset, comprising 150,000 samples, stands out significantly in scale when compared to ExpClip, which consists of 50,000 samples.

\section{Method}

\subsection{Models}
In this section, we propose several baseline models for the task of translating emotion descriptions into 3D facial expressions. This includes \textbf{(i)} fine-tuning of pre-trained language models, \textbf{(ii)} CLIP-based approaches, and \textbf{(iii)} Emotion-XLM, an architecture we have developed to enhance the language model's functionality.

\noindent\textbf{Pretrained LM Baselines:} We utilize BERT \cite{devlin-etal-2019-bert} and Glot500, a highly multilingual variant of XLM-RoBERTa \cite{imanigooghari-etal-2023-glot500}, as the backbones. To map LM outputs into a designated target space, we incorporate a Multi-Layer Perceptron (MLP).
The MLP is trained with tuples $T = \{(b, l) \mid b \in \mathbb{R}^{768}, l \in \mathbb{R}^{52}\}$, where $b$ denotes the LM output and $l$ represents the corresponding blendshape scores.

\noindent\textbf{Emotion-XLM:} Extending the MLP structure to XLM-RoBERTa, we introduce an emotion-extractor unit. The transformer output is fed into this unit to extract emotion distributions alongside one-hotted vectors. Representing the input space as $B =\{b \mid b \in \mathbb{R}^{768}\}$, the emotion-extractor unit produces output $E = \{(v,o) \mid v, o \in \mathbb{R}^{8} \}$, where $v$ indicates emotion intensities in $V = \{[v_1, \ldots , v_8] \mid v_i \in [0,1], i = 1, \ldots, 8\}$, and $o$ is the one-hotted vector of $v$. Pairs of vectors are then passed to the MLP unit, where they are concatenated with the text embedding before being fed to the regression unit, $\mathbb{F}(.) : \mathbb{R}^{784} \to \mathbb{R}^{52}$. In the training time, 50 \% of the times, ground-truth emotion labels are replaced with emotion-extractor unit's output, to efficiently train both modules, ensuring that the blendshape MLP unit is well-trained while giving enough feedbacks to the emotion-extractor unit.

\begin{equation}
\mathcal{L} = \lambda_1\mathcal{L}_{Blendshape} + \lambda_2 \mathcal{L}_{Emotion}
\end{equation}

\begin{figure}[htbp!]
\centering
        {\includegraphics[width=0.45\textwidth]{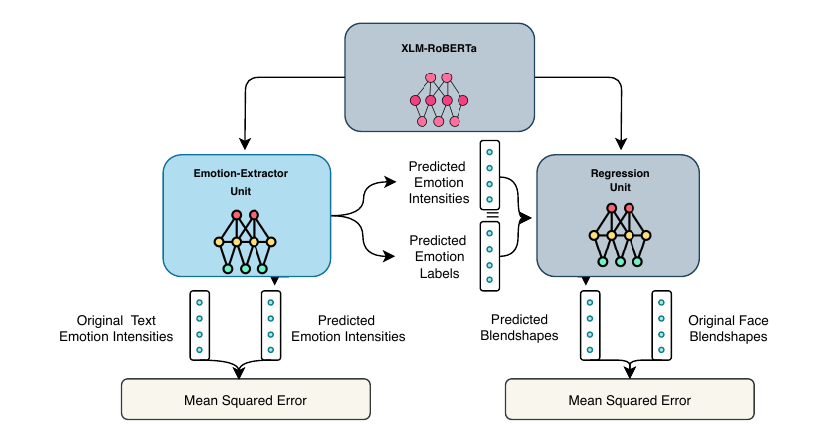}}
        \caption{Emotion-XML uses emotion ground truth to predict facial blendshapes. An Emotion Extractor guides the Regression model with the Teacher-Forcing technique at a 50\% ratio. Both units are trained via Mean Squared Error (MSE) loss. }
        \label{fig:emotion-xlm}  
\end{figure}

Our training methodology employs a combination of MSE losses for blendshapes and extracted emotions, weighted by coefficients to balance their contributions effectively. This model is illustrated in Figure \ref{fig:emotion-xlm}\\

\noindent\textbf{CLIP Baseline:} We employed a Multi-Layer Perceptron (MLP) structure on the CLIP~\cite{Radford2021} backbone. What distinguishes this model from Pretrained LM Baselines is the incorporation of both image and text embeddings during training, effectively doubling the size of our dataset.

\begin{figure}[htbp!]
     \centering
        {\includegraphics[width=0.5\textwidth]{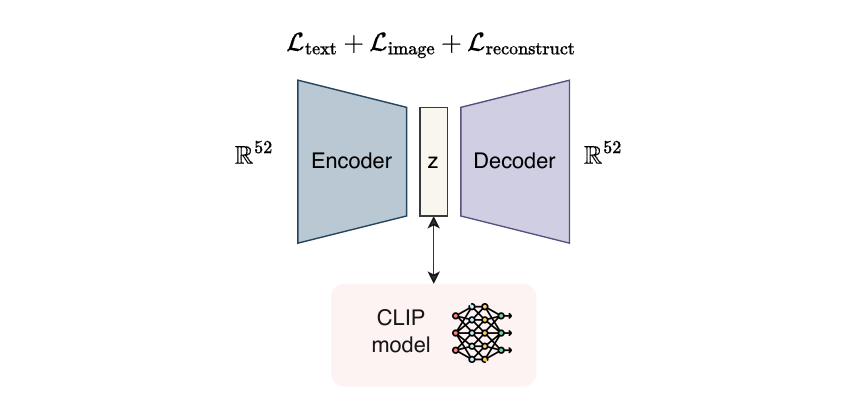}}
        \caption{VAE CLIP concurrently reconstructs facial expressions while aligning their latent representation with corresponding text and image representations in the CLIP space.}
        \label{fig:aeclip}
\end{figure}

\noindent\textbf{VAE CLIP}
We employed a Variational Autoencoder (VAE) to align blendshape scores with their corresponding text and image CLIP~\cite{Radford2021} embeddings, as illustrated in Figure \ref{fig:aeclip}. The encoder maps blendshape scores to their respective text and image representations and the decoder sample from the latent space and reconstructs the blendshape scores. The model is trained via three distinct losses. Textual-blendshape and Visual-blendshape alignment are addressed using cosine similarity. Moreover, The reconstruction loss is defined by Mean Squared Error (MSE).

{\small\begin{align}
&\mathcal{L}_{text} = 1 - \cos{({CLIP}_{text},z)} \\
&\mathcal{L}_{image} = 1 - \cos{({CLIP}_{image},z)} \\
&\mathcal{L} = \mathcal{L}_{recon} + \lambda_{text}\mathcal{L}_{text} + \lambda_{image}\mathcal{L}_{image}
\end{align}}
\\
Here, cos(a, b) denotes the cosine similarity between two vectors a and b.\\

\subsection{Emo3D Metric}
\begin{figure}[htbp!]
     \centering
        {\includegraphics[width=0.45\textwidth, keepaspectratio]{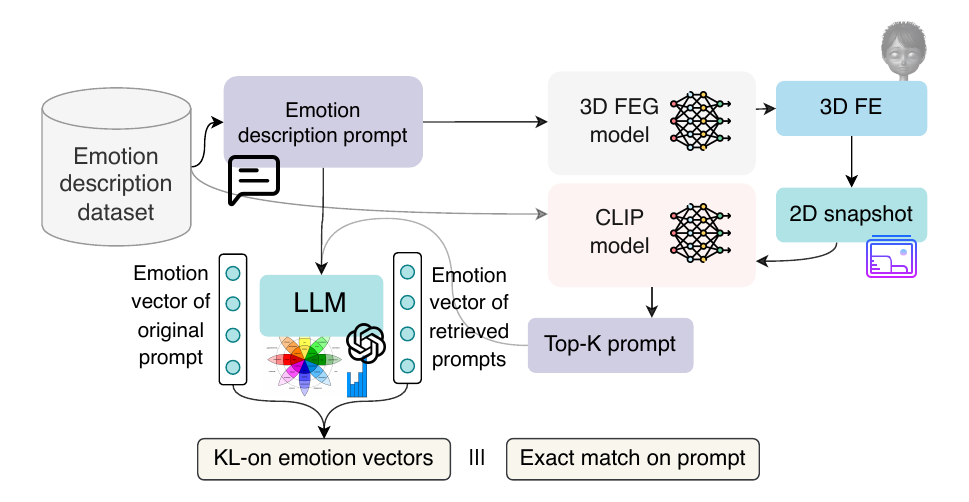}}
        \caption{Our methodology in Emo3D metric entails selecting $n$ prompts with a balanced emotion distribution. We generate facial expressions using a text-utilizing FEG model for a given input prompt. We project the 3D face model onto a 2D image and employ zero-shot CLIP to identify the $k$ nearest text prompts. Subsequently, we compute the Kullback-Leibler (KL) divergence between the emotion distribution of the input text and these $k$ prompts.}
        \label{fig:metric}
\end{figure}

We introduce a new 3D FEG metric for evaluating the reconstruction of the original emotion vector from 2D snapshots of the generated 3D faces. 
We create a test set comprising diverse emotion prompts uniformly selected. To evaluate any proposed FEG model, we generate the corresponding blenshape scores of the input text and project the 3D face model onto a 2D image. Using zero-shot CLIP~\cite{Radford2021}, we identify the k-nearest text prompts related to the image. We calculate the emotion distribution for the original prompt and the top-K prompts. This is followed by computing the Kullback-Leibler (KL) divergence between the emotion vector of the original prompt and the average emotion vector of the top-K retrieved prompts. We refer to the normalized KL bounded between 0 and 1 as the ``Emo3D metric'':
{\small
\begin{equation}
    D_{{KL}}(\phi \, || \, \bar{\psi}) = \sum_{i} \phi(i) \cdot \log\left(\frac{\phi(i)}{\bar{\psi}(i)}\right)
\end{equation}}
{\small
\begin{equation}
    \text{Emo3D Metric} = \frac{1}{1 + e^{-D_{{KL(\phi \, || \, \bar{\psi})}}}}
\end{equation}}

where $\phi$ represents the emotion distribution of the input prompt, and $\bar{\psi}$ represents the mean emotion distribution of the top-k prompts. The steps for Emo3D calculation are outlined in Figure \ref{fig:metric}. In our evaluation of the FEG models, we provide both the Emo3D Metric and the MSE scores of the 3D models for comparison purposes.

\section{Results}
\begin{figure*}[htbp!]
     \centering
        {\includegraphics[]{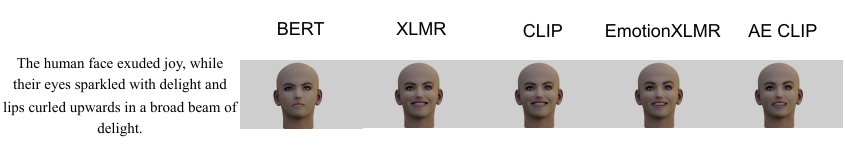}}
        \caption{Model Comparison for the proposed FEG baselines on an example description.}
        \label{fig:comparison}
\end{figure*}

The FEG model performances are provided in Table~\ref{tab:res}. It becomes evident that the CLIP With Regression Unit model demonstrates superior performance when evaluated using our Emo3D metric. 
Our results indicate that the MSE and Emo3D metrics do not consistently align. When we examined the 3D model outputs, we observed that samples that performed better according to Emo3D metric also demonstrated a closer visual resemblance to the input prompt, in contrast to samples that showed better performance based on MSE, similar to Figure \ref{fig:comparison}. This can be because in our metric, Emo3D prioritizes visual-text alignment and proximity, tending to capture richer semantic information than distance metrics in 3D space using MSE.

\begin{table}
  \centering
  \begin{tabular}{@{}l@{ \hskip 0.65in}c@{ \hskip 0.65in}rc@{}}
    \toprule
    Model & MSE & Emo3D\\
    \midrule
    BERT &  0.03   & 0.796\\
    \midrule
    XLMRoBERTa & 0.04 & 0.789\\
    \midrule
    Autoencoder CLIP & 0.002 & 0.776\\
    \midrule
    Emotion-XLM & 0.035 & 0.756\\
    \midrule
    CLIP &  0.014 & 0.737\\
    \bottomrule
  \end{tabular}
  \caption{Performance comparison of FEG models using MSE vs. Emo3D metrics.}
  \label{tab:res}
\end{table}

\section{Conclusion}
In this paper, we introduced ``Emo3D'', a comprehensive ``Text-Image-Expression dataset'' that covered a wide range of human emotions and their textual descriptions, paired with images and 3D blendshapes. Our use of Language Models (LLMs) to generate prompts captured a variety of emotional expressions and descriptions. To the best of our knowledge, ``Emo3D'' stood out as the most comprehensive FEG dataset, encompassing sufficiently diverse and complex emotional descriptions. Furthermore, we developed an efficient evaluation metric to provide 3D image synthesis models with a reliable benchmark. Throughout our work, we tested several unimodal and multimodal models as baselines to encourage new entrants to the field. The significance of ``Emo3D'' lay in its potential to advance 3D facial expression synthesis, holding promising implications for animation, virtual reality, and emotional human-computer interaction.

\section{Limitations and future work}

While our dataset exhibits positive attributes, it is not without errors stemming from the processes involved in its production. Specifically, the use of Mediapipe to obtain blendshape sores introduced inaccuracies, particularly in the representation of certain emotions and facial expressions. To enhance the dataset in future endeavors, collaboration with skilled animators could be sought to refine and design more accurate blendshape scores.

\section{Ethics}
This paper leverages GPT-3.5~\cite{OpenAI2023} for generating textual emotional descriptions and DALL-E3~\cite{ramesh2022hierarchical} for creating corresponding images. It's vital to recognize the potential biases and privacy concerns inherent in these AI models. Both GPT-3.5 and DALL-E3, like many advanced AI systems, reflect the data on which they were trained, which can include societal biases and inaccuracies. Mitigating and analyzing such biases is beyond the scope of this paper. Studies such as ``DALL-EVAL: Probing the Reasoning Skills and Social Biases of Text-to-Image Generation Models'' ~\cite{cho2023dalleval} and the paper ``ChatGPT: A Comprehensive Review on Background, Applications, Key Challenges, Bias, Ethics, Limitations and Future Scope''~\cite{RAY2023121} thoroughly examine biases in the DALL-E and GPT models, respectively. According to these studies, the DALL-E and GPT models are also shown to have certain degrees of biases related to gender, skin tone, professions, and certain attributes, and may have privacy or accountability concerns.


\nocite{*}
\bibliographystyle{icml2021}
\bibliography{main}










 \appendix

 \section{Appendix}

\subsection{Additional Details on Human Evaluation}\label{sec.human}
We recognize that there may be inaccuracies in the emotion dataset, particularly in scenarios involving LLM-based generation. As outlined in the metric section, our methodology primarily employs the distribution of emotions found within the text descriptions. Thus, evaluating the emotions in these text descriptions is deemed sufficient.  In order to do this we introduced a human evaluation component to verify the accuracy of these text-based emotions. Two independent reviewers assessed the emotional content of 100 text samples, with a Cohen's kappa agreement score of 0.88. We then compared these human evaluations with our system's predictions. The results were encouraging, showing an $81\%$ agreement rate between our system and the human evaluators, which indicates a strong correlation.

\subsection{More analysis of Emo3D-dataset}\label{sec.Analysis}
\begin{figure*}[b]
     \centering
        {\includegraphics[width=\textwidth]{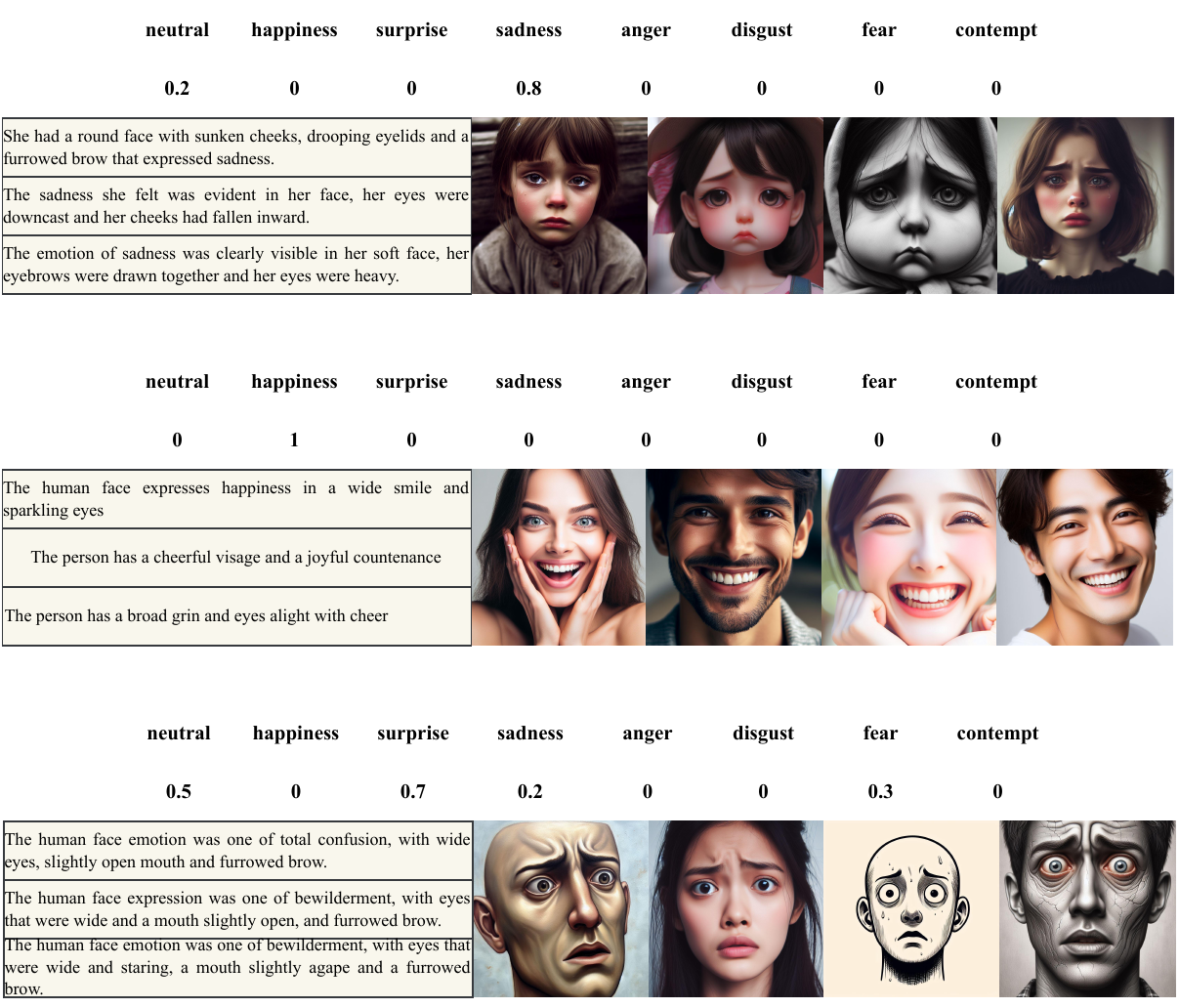}}
        \caption{\textbf{Samples of Emo3d Dataset}: A glimpse into the rich diversity and complexity of our collected data, paving the way for insightful analysis and discovery.}

        \label{fig:data}
\end{figure*}

Figure \ref{fig:data} provides an overview of the dataset. Three sample data points from the dataset are presented in this figure. For each sample, we have three textual descriptions and the corresponding images generated by the DALL-E model \cite{ramesh2022hierarchical}. As can be seen, the generated images align well with the textual descriptions. Additionally, for each sample, we have provided a vector representing the distribution of eight primary senses associated with the given descriptions.

To further explore the linguistic characteristics of each emotion category, this appendix presents three analytical graphs and four detailed tables.

\begin{figure*}[htbp!]
     \centering
        {\includegraphics[width=\textwidth, height = 0.25\textheight]{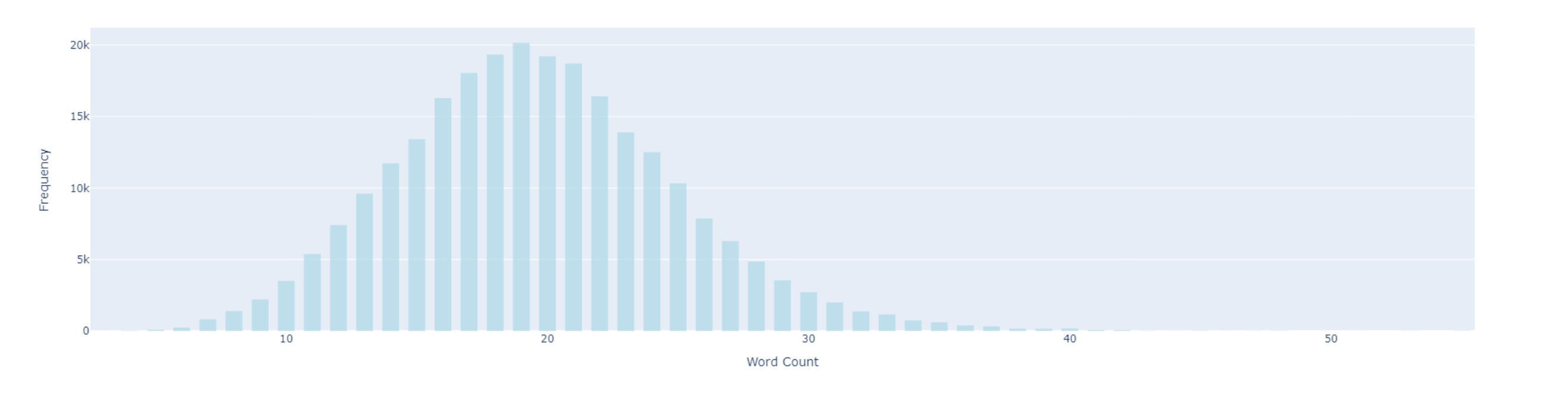}}
        \vspace{-8 mm}

        \caption{\textbf{Distribution of word counts within data}: gives intuition about lengths of textual prompts used within the dataset.}
        \label{fig:WordCounts}
\end{figure*}

\begin{figure*}[htbp!]
     \centering
        {\includegraphics[width=\textwidth]{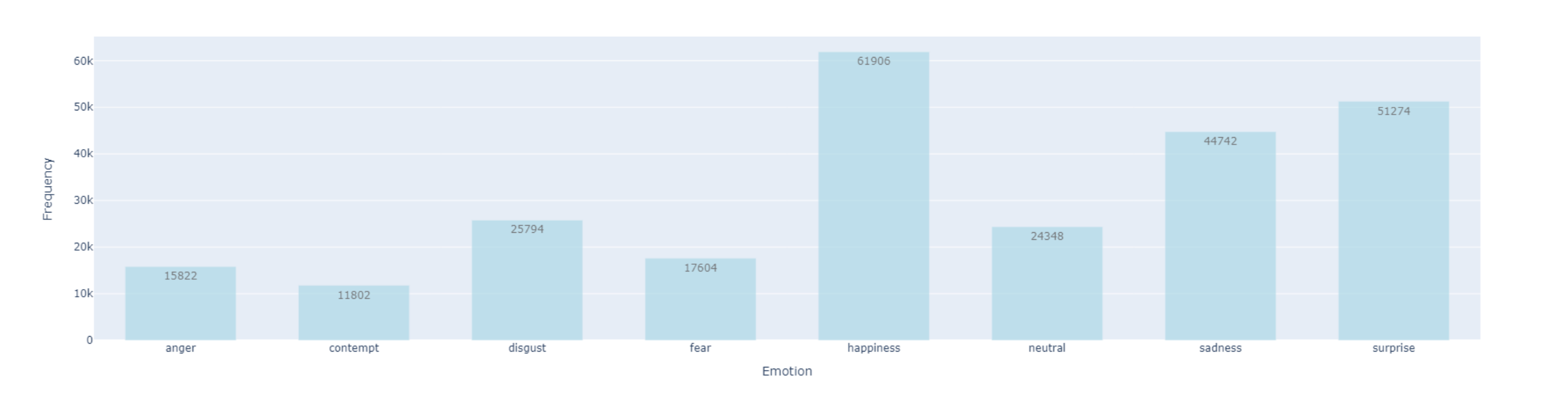}}
        \vspace{-8 mm}
        \caption{Distribution of emotions within prompts:  shows the number of prompts in each emotion class. Note that we count prompt $P$  in emotion class $E$ when the share of $E$ within the emotion distribution of $P$ is the highest.}
        \label{fig:EmotionDist}
\end{figure*}

\begin{figure*}[htbp!]
     \centering
        {\includegraphics[width=\textwidth]{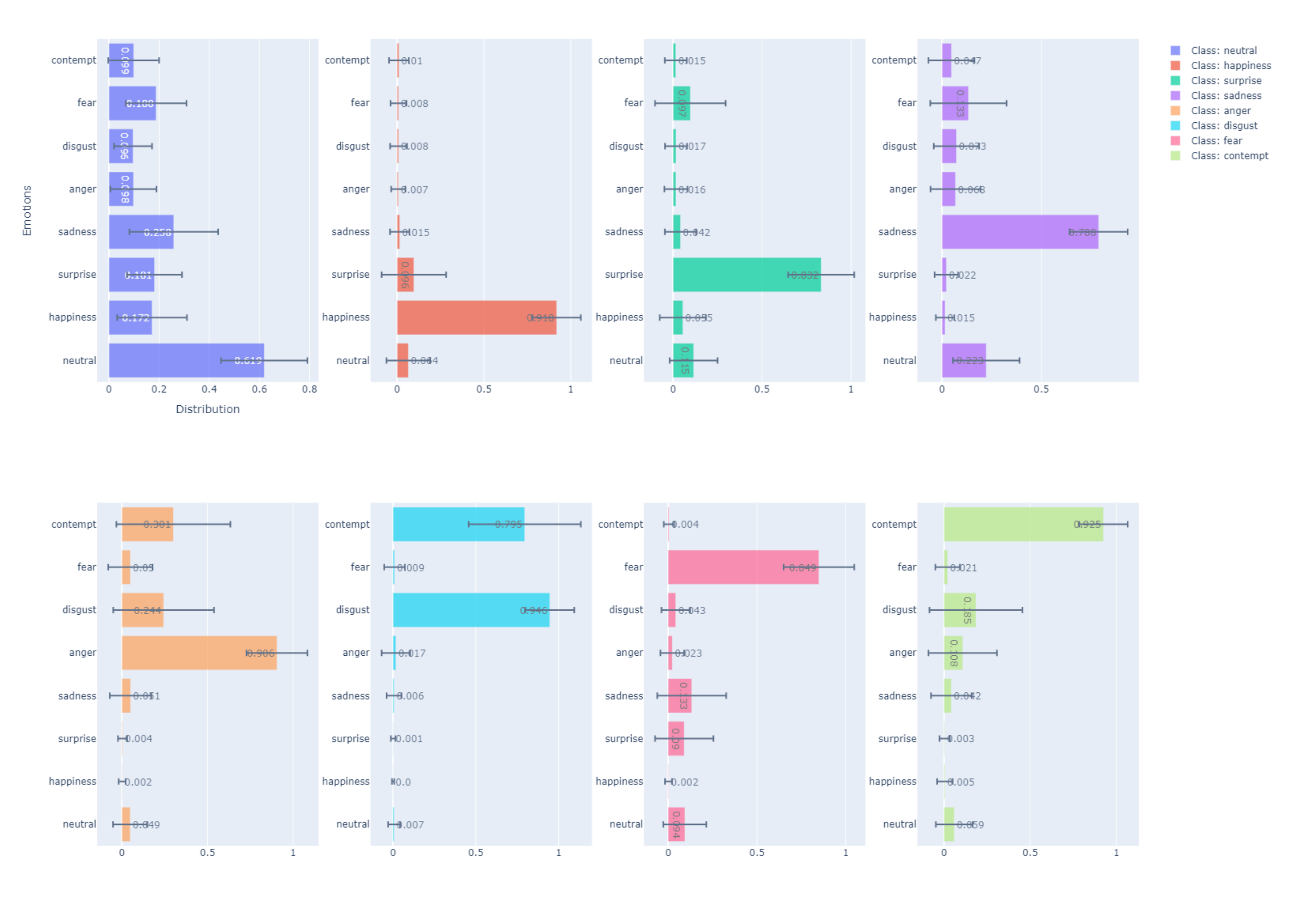}}
        \vspace{-8 mm}

        \caption{Mean \& STD for each class of emotion: shows the mean \& std of each emotion per each major emotion class, i.e. the mean \& STD of emotions within prompts labeled as $E$ where $E$ is an emotion class.}
        \label{fig:MeanSd}
\end{figure*}

Figure \ref{fig:WordCounts} presents the distribution of the number of words in the dataset prompts, providing a general intuition about the length of the text descriptions.\\
Figure \ref{fig:EmotionDist} displays the distribution of different emotions in the dataset. \\
The diagram in Figure \ref{fig:MeanSd} displays the mean and standard deviation of intensity levels for different emotions within each category. By analyzing this figure, we can infer the correlation and simultaneity of different emotions to some extent. For instance, both disgust and contempt show the highest values for each other in both categories. Furthermore, the high values for surprise and sadness in relation to the emotion of fear may suggest a connection between these emotions. This dataset, encompassing various emotion distributions, can be a valuable resource in understanding human emotional experiences.

 Table ~\ref{tab:text analysis table} presents statistical analyses of the dataset, offering further insights into its characteristics. Table ~\ref{tab:frequent words} lists the most frequently occurring words within each category, providing insights into the vocabulary most closely associated with different emotional states. Tables ~\ref{tab:frequent Synset 1-4} and ~\ref{tab:frequent synset 4-8 } delve deeper into the semantic landscape of each emotion by showcasing the most frequent synsets (sets of words with similar meanings) within each category.

\newpage

\begin{table*}[h]  
  \centering
  \begin{tabular}{|c|c|c|c|c|}
    \hline
  Emotion & Number of words &	Number of unique words &	Avg word len &Avg sentence len		\\
    \hline
    Neutral     & 14805	& 1684 & 	7.277609	& 13.555218	    \\
    \hline
    Happiness     & 31405 & 	1519	& 6.687916 & 12.375832    \\
    \hline
    Surprise     & 33690 & 1559 & 6.576492 & 12.152983	    \\
    \hline
    Sadness     & 32878 & 	2220 & 	6.816656 & 	12.633311	    \\
    \hline
    Anger     & 16097 & 1288 & 	6.419271 & 	11.838541    \\
    \hline
    Disgust     & 19917 & 1306 & 	6.766280 & 	12.532560    \\
    \hline
    Fear     & 15120 & 	1245 & 	6.161111 & 	11.322222	    \\
    \hline
    Contempt     & 7535	& 958 & 6.982349 & 	12.964698	    \\
    \hline
    
  \end{tabular}
    \caption{Dataset Statistics by emotion Category}
  \label{tab:text analysis table}
\end{table*}

\begin{table*}[h]  
  \centering
  \begin{tabular}{|c|c|c|c|c|c|c|c|}
    \hline
    Neutral & Happiness & Surprise & Sadness & Anger & Disgust & Fear & Contempt \\
    \hline 
     emotion & happiness & surprise & sadness & anger & contempt & fear & contempt\\
    \hline 
     expression & eyes & eyes & eyes & eyes & expression & eyes & expression\\
    \hline 
     confusion & smile & wide & expression & furrowed & disgust & wide & eyes\\
    \hline 
     person & joy & mouth & downturned & expression & look & mouth & lips\\
    \hline 
     furrowed & wide & emotion & emotion & lips & lips & expression & disdain\\
    \hline 
     one & expression & open & mouth & brow & eyes & furrowed & look\\
    \hline 
     eyes & bright & eyebrows & deep & rage & mouth & lips & emotion\\
    \hline 
     hint & expressing & raised & person & narrowed & furrowed & pale & mouth\\
    \hline 
     random & emotion & shock & lips & eyebrows & nose & open & feeling\\
    \hline 
     look & person & slightly & sorrow & brows & disdain & look & sneer\\
     \hline
        
  \end{tabular}
  \caption{Most Frequent Words for each emotion}
  \label{tab:frequent words}
\end{table*}

\begin{table*}[h]  
  \centering
  \begin{tabular}{|c|c|c|c|}
    \hline
    Neutral & Happiness & Surprise & Sadness \\
    \hline 
     demonstration.n.05 & feeling.n.01 & astonishment.n.01 & area.n.01\\
    \hline 
     cognitive state.n.01 & communication.n.02 & feeling.n.01 & feeling.n.01\\
    \hline 
     combination.n.07 & area.n.01 & demonstration.n.05 & sadness.n.01\\
    \hline 
     confusion.n.02 & positive stimulus.n.01 & cognitive state.n.01 & negative stimulus.n.01\\
    \hline 
     communication.n.02 & demonstration.n.05 & combination.n.07 & region.n.01\\
    \hline 
     feeling.n.01 & collection.n.01 & emotion.n.01 & sagacity.n.01\\
    \hline 
     countenance.n.01 & emotional state.n.01 & communication.n.02 & unhappiness.n.02\\
    \hline 
     sagacity.n.01 & sagacity.n.01 & sagacity.n.01 & countenance.n.01\\
    \hline 
     small indefinite quantity.n.01 & facial expression.n.01 & rejoinder.n.01 & communication.n.01\\
    \hline 
     hair.n.01 & countenance.n.01 & hair.n.01 & rejoinder.n.01\\
     \hline
        
  \end{tabular}
  \caption{Most frequent synsets for each emotion}
  \label{tab:frequent Synset 1-4}
\end{table*}

\begin{table*}[h]  
  \centering
  \begin{tabular}{|c|c|c|c|}
    \hline
    Anger & Disgust & Fear & Contempt \\
    \hline 
     feature.n.02 & dislike.n.02 & fear.n.01 & dislike.n.02\\
    \hline 
     anger.n.01 & demonstration.n.05 & feature.n.02 & demonstration.n.05\\
    \hline 
     communication.n.02 & area.n.01 & emotion.n.01 & area.n.01\\
    \hline 
     sagacity.n.01 & facial expression.n.01 & sagacity.n.01 & hair.n.01\\
    \hline 
     countenance.n.01 & communication.n.02 & anxiety.n.02 & communication.n.02\\
    \hline 
     demonstration.n.05 & region.n.01 & countenance.n.01 & region.n.01\\
    \hline 
     communication.n.01 & rejoinder.n.01 & communication.n.01 & disrespect.n.01\\
    \hline 
     hair.n.01 & countenance.n.01 & rejoinder.n.01 & countenance.n.01\\
    \hline 
     rejoinder.n.01 & communication.n.01 & hair.n.01 & communication.n.01\\
    \hline 
     feeling.n.01 & disrespect.n.01 & appearance.n.01 & rejoinder.n.01\\
     \hline
  \end{tabular}
  \caption{Most Frequent synsets  for each emotion}
  \label{tab:frequent synset 4-8 }
\end{table*}

\end{document}